\documentclass[10pt,twocolumn,letterpaper]{article}

\usepackage[pagenumbers]{cvpr} 

\usepackage{graphicx}
\usepackage{amsmath}
\usepackage{amssymb}
\usepackage{booktabs}

\usepackage[pagebackref,breaklinks,colorlinks]{hyperref}

\usepackage[capitalize]{cleveref}
\crefname{section}{Sec.}{Secs.}
\Crefname{section}{Section}{Sections}
\Crefname{table}{Table}{Tables}
\crefname{table}{Tab.}{Tabs.}

\def\cvprPaperID{*****} 
\def\confName{CVPR}
\def\confYear{2022}

\begin{document}

\title{Phy124: Fast Physics-Driven 4D Content Generation from a Single Image}

\author{%
    Jiajing Lin \\
    School of Informatics\\
    Xiamen University\\
    \texttt{31520231154298@stu.xmu.edu.cn} \\
    \and
    Zhenzhong Wang \\
    Department of Computing\\
    The Hong Kong Polytechnic University\\
    \texttt{zhenzhong16.wang@connect.polyu.hk} \\
    \and
    Yongjie Hou \\  
    School of Informatics\\
    Xiamen University\\
    \texttt{23120231150268@stu.xmu.edu.cn} \\
    \and
    Yuzhou Tang \\
    School of Informatics\\
    Xiamen University\\
    \texttt{juliantang@stu.xmu.edu.cn} \\
    \and
    Min Jiang\thanks{Corresponding author} \\
    School of Informatics\\
    Xiamen University\\
    \texttt{minjiang@xmu.edu.cn} \\
}
\maketitle

\begin{abstract}
4D content generation focuses on creating dynamic 3D objects that change over time. Existing methods primarily rely on pre-trained video diffusion models, utilizing sampling processes or reference videos. However, these approaches face significant challenges. Firstly, the generated 4D content often fails to adhere to real-world physics since video diffusion models do not incorporate physical priors. Secondly, the extensive sampling process and the large number of parameters in diffusion models result in exceedingly time-consuming generation processes. To address these issues, we introduce \textit{Phy124}, a novel, fast, and physics-driven method for controllable 4D content generation from a single image. \textit{Phy124} integrates physical simulation directly into the 4D generation process, ensuring that the resulting 4D content adheres to natural physical laws. \textit{Phy124} also eliminates the use of diffusion models during the 4D dynamics generation phase, significantly speeding up the process. \textit{Phy124} allows for the control of 4D dynamics, including movement speed and direction, by manipulating external forces. Extensive experiments demonstrate that \textit{Phy124} generates high-fidelity 4D content with significantly reduced inference times, achieving state-of-the-art performance. The code and generated 4D content are available at the provided link: \url{https://anonymous.4open.science/r/BBF2/}.
\end{abstract}

\section{Introduction}
\label{sec:intro}
The generation of 4D content is becoming increasingly valuable in a variety of fields, such as animation, gaming, and virtual reality~\cite{Art_Create}. Recent advancements in diffusion models~\cite{DM} have revolutionized image generation~\cite{DM-image-1, DM-image-2} and video generation~\cite{DM-video-1, DM-video-2}. These models' robust visual priors have significantly propelled progress in 4D content generation~\cite{Animate124, 4D-FY}. Such progress has made the automated production of high-quality 4D content not only achievable but also increasingly efficient.

\begin{figure}[!t]
	\centering
	\includegraphics[width=1\linewidth]{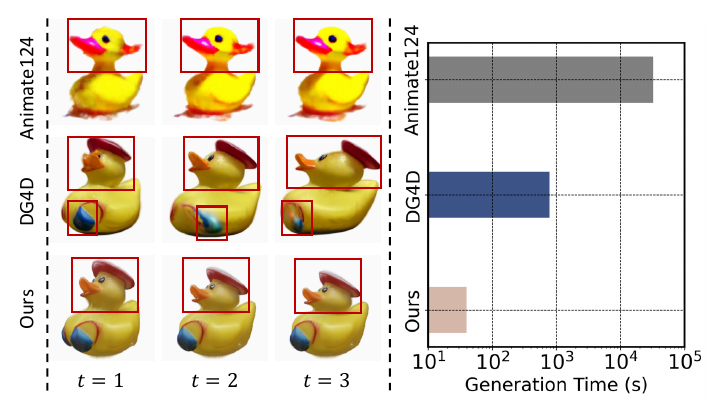}
	\caption{The figure shows a duck being pressed. Animate124 struggles to produce effective motion and often results in abnormal appearances in the 4D content; for instance, the duck in the figure has two beaks. On the other hand, DG4D produces motion that does not adhere to physical laws, such as the abnormal deformations observed in the duck's head and blue wing.
    In contrast, our approach can generate 4D content with greater physical accuracy. Moreover, our method reduces the generation time to an average of just 39.5 seconds.}
	\label{fig:intro}
\end{figure}

Generating 4D content from a single image presents a formidable challenge due to the inherent absence of both temporal and spatial information in a single image. To overcome this limitation, existing methods~\cite{Animate124, DreamGaussian4D} predominantly rely on pre-trained video diffusion models to capture dynamic information. These methods are generally categorized into two primary approaches: temporal score distillation sampling (SDS)-driven and reference video-driven.

In the temporal SDS-driven category, Animate124~\cite{Animate124} is a pioneering framework that adopts a coarse-to-fine strategy for image-to-4D generation. Initially, 2D video diffusion and 3D-aware diffusion~\cite{Zero123} are employed for rough optimization, followed by detailed refinement using a ControlNet prior~\cite{DM-image-2}. However, Animate124 faces significant limitations, including difficulties in generating 4D content that accurately corresponds to the input image and the requirement for extensive generation time.
Alternatively, reference video-driven methods have been developed to enhance both generation efficiency and output quality. DreamGaussian4D~\cite{DreamGaussian4D} employs deformable 3D Gaussians~\cite{Deformable3DGS} as 4D representations, in which a 3D-aware diffusion model refines the 4D representations to match the rendered video to a reference video.

Despite these advancements, existing methods still face several critical challenges. 
Firstly, the reliance on video diffusion models, which often fail to effectively learn physical principles, results in 4D content that frequently violates physical laws, as illustrated in the left part of Fig. \ref{fig:intro}. Secondly, the large scale of pre-trained image or video diffusion models, coupled with the frequent use of SDS, leads to excessively time-consuming generation processes, as clearly shown in the right part of Fig. \ref{fig:intro}. Additionally, the inherently stochastic nature of video diffusion models often results in uncontrollable dynamics within the generated 4D content.

In this work, we introduce \textit{Phy124}, a physics-driven method capable of generating controllable 4D content from a single image in just 39.5 seconds, as demonstrated in Fig. \ref{fig:intro}. Our key innovation is the integration of physical simulation into the 4D generation process. Unlike most current methods~\cite{Animate124, DreamGaussian4D, Diffusion4D, EG4D} that rely on video diffusion models to generate dynamics in 4D content, \textit{Phy124} applies physical simulation to ensure that the generated 4D content adheres to physical laws. Moreover, by eliminating the time-consuming SDS during the 4D dynamics generation phase, \textit{Phy124} achieves rapid 4D content generation from a single image. To provide precise control over the dynamics in the generated 4D content, we introduce external forces. By fine-tuning these external forces, \textit{Phy124} can generate 4D content that aligns with user intent. The main contributions of our work are summarized as follows:

\begin{itemize}
    \item We present a physics-driven image-to-4D generation framework that integrates physical simulation directly into the 4D generation process, enabling the rapid production of 4D content that adheres to physical principles.
    \item We first introduce external forces to facilitate the generation of controllable 4D content, allowing for precise control over the dynamics in the generated 4D content.
\end{itemize}

Qualitative and quantitative comparisons demonstrate that our method generates 4D content that is physically accurate, spatially and temporally consistent, high-fidelity, and controllable, with significantly reduced generation times.
\section{Related Work}
\subsection{3D Generation.} 
The development of 2D vision~\cite{gong2021eliminate,gong2022person,gong2024beyond,gong2024beyond2,gong2024cross,gong2024exploring} has greatly inspired the advancement of 3D vision.
The objective of 3D generation is to produce 3D content that is consistent with the given conditions, such as text descriptions and reference images. DreamFusion ~\cite{DreamFusion} first introduced SDS, a method for distilling geometric and appearance information from diffusion models to optimize 3D models.
Subsequent image-to-3D~\cite{3DGeneration-image-1,3DGeneration-image-2,3DGeneration-image-3} and text-to-3D~\cite{3DGeneration-text-1,3DGeneration-text-2,3DGeneration-text-3, 3DGeneration-text-4, 3DGeneration-text-5} based on the SDS paradigm has emerged, aiming to address issues such as multi-face Janus issues and slow optimization. However, methods based on the SDS paradigm often require considerable optimization time. On the other hand, some studies ~\cite{Zero123, Wonder3d, SyncDreamer, One2345} focus on generating spatially consistent multi-view images for subsequent 3D reconstruction. However, achieving better multi-view consistency to enhance the quality of the generated 3D Gaussians remains an open problem. Additionally, some research ~\cite{Shape-e, Point-e} trains models to directly generate 3D point clouds or meshes. However, obtaining 3D point cloud data or meshes is expensive.
\subsection{4D Generation.} 4D generation ~\cite{Nvidia, Efficient4D, Gaussianflow} aims to generate dynamic 3D content that aligns with input conditions such as text, images, and videos. Currently, most 4D generation work ~\cite{4D-FY, Consistent4D, Animate124, 4DGen, Dreamscene4d} heavily depends on diffusion models~\cite{DM-implicit}. 
Based on the input conditions, 4D generation can be categorized into three types: text-to-4D, video-to-4D, and image-to-4D. For instance, MAV3D ~\cite{MAV3D} is the first text-to-4D work that trains HexPlane~\cite{Hexplane} using temporal SDS from a text-to-video diffusion model. 4D-fy ~\cite{4D-FY} tackles the text-to-4D task by integrating various diffusion priors. However, these methods all suffer from time-consuming optimization. Instead of text input, Consistent4D ~\cite{Consistent4D} is an approach for generating 360-degree dynamic objects from monocular video by introducing cascade DyNeRF and interpolation-driven consistency loss. 4DGen~\cite{4DGen} supervised using pseudo labels generated by a multi-view diffusion model. In practice, acquiring high-quality reference videos can be challenging. Animate124~\cite{Animate124} pioneered an image-to-4D framework using a coarse-to-fine strategy that combines different diffusion priors. However, Animate124 struggles to generate 4D content that accurately matches the input image (see Fig. \ref{fig:intro}). The process of generating 4D content from an image in DreamGaussian4D ~\cite{DreamGaussian4D} avoids using temporal SDS and instead performs optimization based on reference videos generated by a video diffusion model. Yet, generating high-quality reference videos with video diffusion models is often challenging. Compared to previous works, our framework can efficiently generate 4D content from a single image, avoiding the time-consuming optimization process during 4D dynamics generation phase, and enabling high-fidelity and controllable 4D generation.


\begin{figure*}[htbp]
	\centering
	\includegraphics[width=1\linewidth]{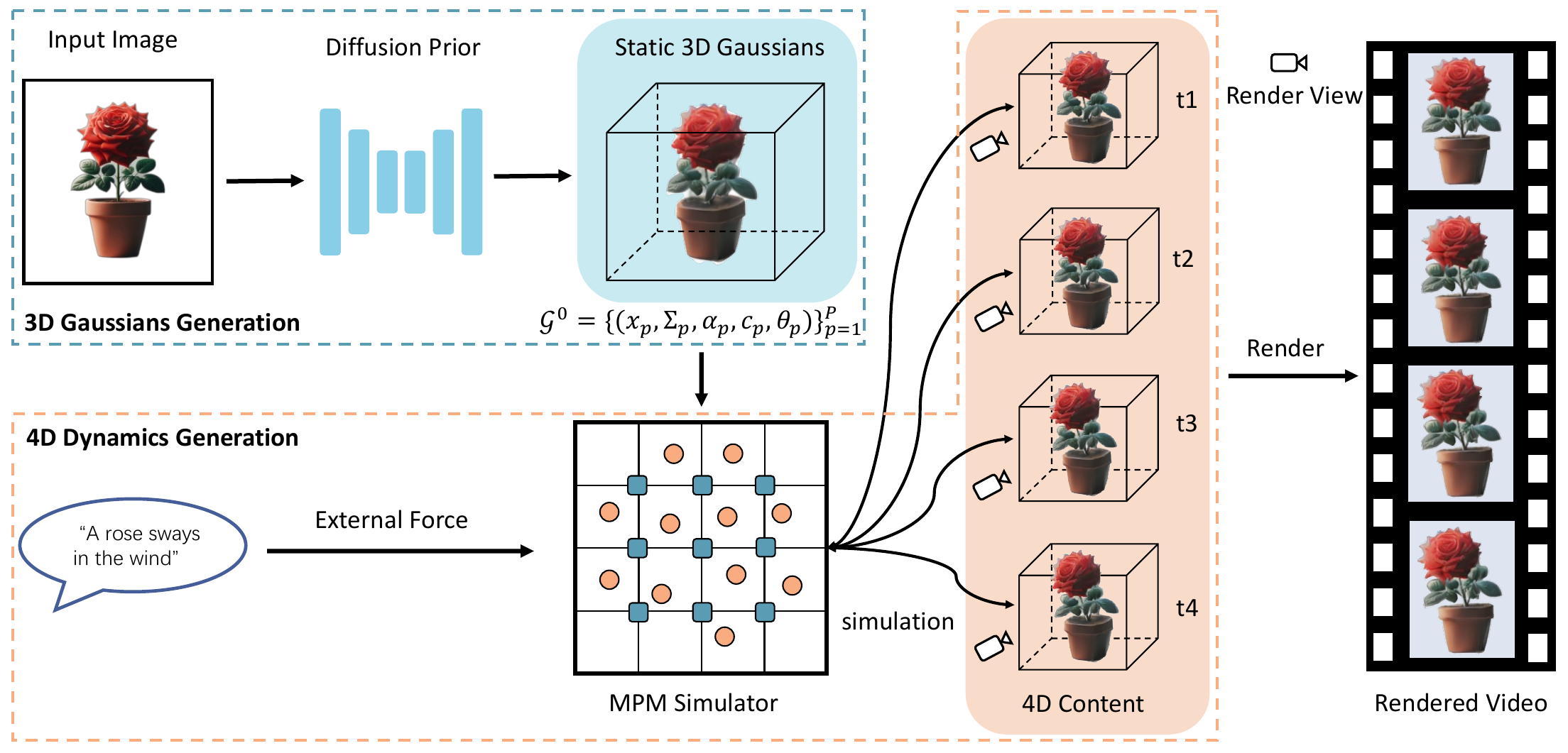}
	\caption{\textbf{Framework of \textit{Phy124}}. 
 In \textbf{3D Gaussians Generation} stage, from an input image, a static 3D Gaussians will be generated under the guidance of the diffusion model. In \textbf{4D Dynamics Generation} stage, we consider each 3D Gaussian kernel as particles within a continuum and attribute physical properties (e.g., density, mass, etc) to them. Sequentially, we employ MPM to introduce dynamics to the static 3D Gaussians. Meanwhile, users can guide the MPM simulator to generate 4D content that aligns with their desired outcomes by adjusting the external forces.}
	\label{fig:pipeline}
\end{figure*}

\section{Preliminaries}
\subsection{3D Gaussian Splatting}  
Unlike NeRF ~\cite{NeRF, Mip-NeRF, Instant-NeRF}, which uses implicit representations, 3D Gaussian Splatting (3DGS) employs explicit representations through a set of anisotropic Gaussian kernels~\cite{3DGS}, which can be interpreted as particles in space, enables physical simulations to simulate the particles' motion. The explicit representations also enable superior reconstruction quality and faster rendering speed. Specifically, 3DGS represents the 3D scene using a set of anisotropic Gaussian kernels defined by center position (mean) $\mathbf{x} \in \mathbb{R}^3$, covariance matrix $\mathbf{\Sigma} \in \mathbb{R}^7$, opacity $\alpha \in \mathbb{R}$, and color $\mathbf{c} \in \mathbb{R}^{3}$.
The 3D Gaussian kernel's spatial distribution can be expressed by:
\begin{equation}
    G(\mathbf{x}) = e^{-{\cfrac{1}{2}}(\mathbf{x})^T \mathbf{\Sigma}^{-1} (\mathbf{x})}.
\end{equation}
During the rendering process, the 3D Gaussians will be projected onto the image plane as 2D Gaussians. Sequentially, the pixel color $\mathbf{C}(\mathbf{\mathbf{r}})$ rendered from ray $\mathbf{r}$ can be computed by point-based volume rendering technique:
\begin{equation}
    \mathbf{C}(\mathbf{r}) = \sum_{i \in \mathcal{N}}\mathbf{c}_i\sigma_i\prod_{j=1}^i(1-\sigma_j),
\end{equation}
where $\sigma_i = \alpha_iG(\mathbf{x}_i)$, and $\mathcal{N}$ denotes the number of Gaussian kernel along ray $\mathbf{r}$. 
Because of this explicit representation, tracking the state of the Gaussian kernels becomes straightforward. In this study, we extend this explicit nature to propose a physics-driven 4D generation framework.
\subsubsection{External Forces}
In physical simulations, external forces directly impact the system's physical behavior. 
External forces are influences acting on a continuum that directly affect its motion and deformation, such as gravity. 
In this paper, all external forces, except for gravity, are applied to the Lagrangian particles which are discrete points that follow the motion of the material, and we apply two types of external forces in total. The first type directly modifies the particle's velocity. For instance, we can set the velocity of all particles to simulate the translation of the continuum. The second type indirectly affects the particle's velocity by applying external forces $\mathbf{f}$. For example, we can simulate the continuum falling from the air by applying gravity. We then use Newton's second law and time integration to compute the particle's velocity at the next time step:
\begin{equation}
    \mathbf{a}_p^{t} = \frac{\mathbf{f}}{m_p^t},
    \quad
    \mathbf{v}_p^{t+1} = \mathbf{v}_p^{t} + \mathbf{a}_p^t \Delta t,
\end{equation}
where $\mathbf{a}_p^t$ denotes the acceleration of particle $p$ at time step $t$, and $\Delta t$ denotes the time interval between time step $t$ and $t+1$.
By adjusting the external forces, we can control the motion and deformation of the object.
\section{Methodology}
 

The proposed \textit{Phy124} is illustrated in Fig \ref{fig:pipeline}. The framework includes two stages: \textbf{3D Gaussians Generation} and \textbf{4D Dynamics Generation}. In \textbf{3D Gaussians Generation}, a static 3D Gaussian is generated from a single image under the guidance of the image diffusion model. Subsequently, in \textbf{4D Dynamics Generation}, dynamics are generated from static 3D Gaussians by integrating physical simulation, i.e., material point method (MPM) simulator. Particularly, MPM allows for the control of 4D dynamics, including movement speed and direction, by manipulating external forces. In the following parts, we show the details of \textbf{3D Gaussians Generation} and \textbf{4D Dynamics Generation}.

\subsection{3D Gaussians Generation}
Due to the high cost of 3D data acquisition, training a 3D generator directly from 3D data is challenging. Therefore, many works attempt to lift pre-trained 2D diffusion models to guide 3D Generation. DreamFusion~\cite{DreamFusion} first introduced SDS. In this paradigm, the 3D scene is represented as $\vartheta$, which can be either NeRF ~\cite{NeRF}, or 3DGS ~\cite{3DGS} and optimized by SDS. Given a random camera pose, an image can be rendered as $\mathbf{I}=g(\vartheta)$ where $g$ denotes the differentiable rendering process. During the 3D generation process, $\vartheta$ is iteratively optimized to ensure that images rendered from any given view conform to the distribution of the diffusion model. The optimization loss can be expressed as:
\begin{equation}
\nabla_{\vartheta} \mathcal{L}_{\text{SDS}} = \mathbb{E}_{\epsilon, t} \left[ w(t) \left( \hat{\epsilon}_{\phi}(\mathbf{I}_t; y, t) - \epsilon \right) \frac{\partial \mathbf{I}}{\partial \vartheta} \right],
\end{equation}
where $\epsilon$ is the random Gaussian noise added during the diffusion process, $\hat{\epsilon}_\phi$ is the predicted noise for a given noise image $\mathbf{I}_t$ during the denoising process, $y$ represents the conditional information, and $w(t)$ is weighting function. 

However, 3D generation based on score distillation sampling requires a significant amount of time. Recently, some works~\cite{Zero123, MVDream, ImageDream} have explored training 3D-aware diffusion models using 3D data to generate multi-view consistent images for guiding 3D generation.

For 3D Gaussians generation, we can generate static 3D Gaussians from a single image using any image-to-3D Generation method guided by diffusion models, providing plug-and-play capability. Meanwhile, the quality of the 4D content will improve with the quality of the static 3D Gaussians generated from the input image. In this phase, we will obtain a static 3D Gaussians $\boldsymbol{\mathcal{G}}^0 = \{(\mathbf{x}_p, \mathbf{\Sigma}_p, \alpha_p, \mathbf{c}_p)\}_{p=1}^P$ for subsequent simulation, where $\mathbf{x}_p$, $\mathbf{\Sigma}_p$, $\alpha_p$, and $\mathbf{c}_p$ represent the center position, covariance matrix, opacity, and color of the Gaussian kernel $p$, respectively, and $P$ denotes the total number of Gaussian kernels in the 3D Gaussians.

\subsection{4D Dynamics Generation}

In this work, we use the MPM as an engine to drive the dynamics generation of physics. In this part, we briefly introduce MPM. Then, we present the details of how MPM is integrated into our framework.

\subsubsection{Material Point Method.}
Continuum mechanics studies the deformation and motion behavior of materials under forces. 
Motion is typically represented by the deformation map $\mathbf{x} = \phi(\mathbf{X}, t)$,
which maps from the undeformed material space $\omega^0$ to the deformed world space $\omega^t$.
The deformation gradient $\mathbf{F} = \frac{\partial \phi}{\partial \mathbf{X}} (\mathbf{X}, t)$ describes how the material deforms locally. 
MPM is a simulation method that combines Lagrangian particles with Eulerian grids and has demonstrated its ability ~\cite{MPM-1, MPM-3, MPM-4} to simulate various materials. 
In MPM, each particle $p$ carries various physical properties $\boldsymbol{\theta}_p^t$ at time step $t$, including mass $m_p$, density $\rho_p$, volume $V_p$, Young's modulus $E_p$, Poisson's ratio $\nu_p$, velocity $\mathbf{v}_p^t$, deformation gradient $\mathbf{F}_p^t$ and velocity gradient $\mathbf{C}_p^t$. The grid $i$ is used for computing intermediate results.
MPM operates within a loop that includes particle-to-grid (P2G) transfer, grid operations, and grid-to-particle (G2P) transfer.
In the particle-to-grid (P2G) stage, MPM transfers momentum and mass from particles to grids:
\begin{align}
(m\mathbf{v})_i^{t+1} &= \sum_p w_{ip} \left[ m_p \mathbf{v}_p^t + m_p \mathbf{C}_p^t (\mathbf{x}_i - \mathbf{x}_p^t) \right], \\
m_i^{t+1} &= \sum_p w_{ip}m_p,
\end{align}
where $w_{ip}$ is the B-spline kernel that measures the distance between particle $p$ and grid $i$. After P2G stage, we perform grid operations:
\begin{equation}
    \hat{\mathbf{v}}_i^{t+1} = (m\mathbf{v})_i^{t+1} / m_i^{t+1}, 
    \quad
    \mathbf{v}_i^{t+1} = \text{BC}(\hat{\mathbf{v}_i^{t+1}}),
\end{equation}
where BC refers to boundary condition. Then we transfer the results back to particles in the grid-to-particle (G2P) stage:
\begin{align}
\mathbf{v}_p^{t+1} &= \sum_i w_{ip} \mathbf{v}_i^{n+1}, \\
\mathbf{x}_p^{t+1} &= \mathbf{x}_p^t + \Delta t \mathbf{v}_p^{t+1}.
\end{align}
The velocity $\mathbf{v}_p^{t+1}$ and position $\mathbf{x}_p^{t+1}$ are updated using semi-implicit Euler method. Then, we update velocity gradient $\mathbf{C}_p^{t+1}$ and deformation gradient $\mathbf{F}_p^{t+1}$:
\begin{align}
\mathbf{C}_p^{t+1} &= \frac{4}{\Delta \mathbf{x}^2} \sum_i w_{ip} \mathbf{v}_i^{t+1} (\mathbf{x}_i - \mathbf{x}_p^t)^T, \\
\mathbf{F}_p^{t+1} &= \left( \mathbf{I} + \Delta t \mathbf{C}_p^{t+1} \right) \mathbf{F}_p^t. 
\end{align}
By following these three stages, we complete a simulation step.
For a more detailed introduction to MPM, refer to ~\cite{MPM-1, MPM-2}.
In this work, we employ MPM to achieve image-to-4D generation.

\subsubsection{Dynamics Generation.} In our work, dynamics in 4D content are generated through physical simulation --- MPM. Particularly, 3D Gaussians can be viewed as a discretization of the continuum, making it straightforward to integrate MPM. Therefore, we treat each Gaussian kernel as a particle in the continuum and endow each Gaussian kernel with time property $t$ and physical properties $\theta$, Therefore, each Gaussian kernel can be expressed as:
\begin{equation}
    \boldsymbol{\mathcal{G}}_p^t = (\mathbf{x}_p^t, \mathbf{\Sigma}_p^t, \sigma_p, \mathbf{c}_p, \boldsymbol{\theta}_p^t).
\end{equation}
Here, the subscript $p$ denotes a specific Gaussian kernel within the continuum.
Following ~\cite{Physgaussian}, we employ MPM to perform physical simulations on the continuum represented by 3D Gaussian Splatting (3DGS). This allows us to track the position and shape changes of each Gaussian kernel at every time step:
\begin{equation}
    \mathbf{x}^{t+1}, \mathbf{F}^{t+1}, \mathbf{v}^{t+1} = \text{MPMSimulation}(\mathcal{G}^{t}),
\end{equation}
where $\mathbf{x}^{t+1} = \{\mathbf{x}_p^{t+1}\}_{p=1}^P$ denotes the positions of all Gaussian kernel at time step $t+1$.
$\mathbf{F}^{t+1} = \{\mathbf{F}_p^{t+1}\}_{p=1}^P$ represents deformation gradients, which describe the local deformation of each Gaussian kernel at time step $t+1$. 
Intuitively, we can consider the deformation gradient as a local affine transformation applied to the Gaussian kernel. Consequently, we can derive the covariance matrix of Gaussian kernel $p$ in step $t+1$:
\begin{equation}
    \mathbf{\Sigma}_p^{t+1} = (\mathbf{F}_p^{t+1}) \mathbf{\Sigma}_p^t (\mathbf{F}_p^{t+1})^T.
\end{equation}
At each step of the MPM simulation, we obtain the motion and deformation information of the continuum represented by static 3D Gaussians.
This enables us to generate 4D dynamics that are consistent with physical constraints. At this phase, we have entirely eliminated the need for diffusion models, which significantly accelerates the time required to produce 4D content.
In particular, by controlling external forces, we can precisely manage the dynamic effects of 4D content generated from a single image. This makes our method more controllable compared to existing approaches ~\cite{Animate124, DreamGaussian4D, Diffusion4D}, which rely on diffusion models and face challenges in precisely controlling the dynamics of 4D content.

\subsubsection{4D Representation.} Because physical simulations consume a certain amount of time, we need to avoid performing simulations each time we render video for a given view. Therefore, we need a 4D representation method to accelerate rendering time after the simulation is complete. We store the 3D Gaussians at each time step, which collectively form a 4D representation. However, at each simulation time step, we can only retrieve the deformation gradient, which is applied to the covariance matrix to represent the deformation of the Gaussian kernel. To store the 3D Gaussians at each time step for efficient rendering, we need to determine rotation quaternion and scale factor. To achieve this, we perform eigenvalue decomposition on the deformed covariance matrix:
\begin{equation}
    \mathbf{\Sigma} = \mathbf{Q} \mathbf{\Lambda} \mathbf{Q}^T = \mathbf{Q} (\mathbf{\Lambda}^{\frac{1}{2}})(\mathbf{\Lambda}^{\frac{1}{2}})^T \mathbf{Q}^T = \mathbf{R}\mathbf{S}\mathbf{S}^T\mathbf{R},
\end{equation}
$\mathbf{\Lambda}$ is a diagonal matrix consisting of eigenvalues, while $\mathbf{Q}$ is a matrix  comprised of eigenvectors. $\mathbf{R}$ and $\mathbf{S}$ represent the rotation and scaling matrices, respectively.
Additionally, we need to convert the rotation matrix into a rotation quaternion for storage. For the specific formulas related to the conversion, please refer to the appendix. 
By following the steps outlined above, the 3D Gaussians for each time step can be saved, forming a 4D representation. 

\section{Experiments}
Our experiment consists of the following parts:
i) Qualitatively and quantitatively demonstrating that our method outperforms state-of-the-art methods;
ii) Verifying the ability to control 4D dynamics;
iii) Showcasing the plug-and-play capability of our approach.

\begin{table}[h!]
\centering
\begin{tabular}{cccc}
\hline
\textbf{Method} & \textbf{Animate124} & \textbf{DG4D} & \textbf{Ours} \\ \hline
CLIP-T-$f$ $\uparrow$ & 0.9931 & 0.9858 & \textbf{0.9962} \\ 
CLIP-T-$r$ $\uparrow$ & 0.9928 & 0.9823 & \textbf{0.9948} \\
CLIP-T-$b$ $\uparrow$ & 0.9880 & 0.9867 & \textbf{0.9960} \\ 
CLIP-T-$l$ $\uparrow$ & 0.9908 & 0.9844 & \textbf{0.9963} \\ 
Time $\downarrow$ & 32433.63s &  807.07s & \textbf{23.89s+15.67s} \\ \hline
\end{tabular}
\caption{Quantitative comparison with other methods in terms of video quality and generation time. 
The notations $f$, $r$, $b$, and $l$ denote the front, right, back, and left rendered views, respectively.
The total generation time for our method is divided into two stages: the static 3D generation phase and the 4D dynamic generation phase.}
\label{tab:main_exp}
\end{table}

\subsection{Experimental Setup}


\textit{Phy124} is compared with current state-of-the-art (SOTA) approaches, including the NeRF-based Animate124 ~\cite{Animate124} and the 3DGS-based DreamGaussian4D ~\cite{DreamGaussian4D}. Following \cite{Animate124}, we conducted qualitative and quantitative evaluations to demonstrate the effectiveness of our method. For the quantitative evaluation, we used CLIP-T scores. Datasets were collected from ~\cite{Zero123, One2345, Animate124, Consistent4D}.
All experiments are performed on NVIDIA A40(48GB) GPU. 
For more detailed information on experimental settings, please refer to the appendix.

The settings for \textit{Phy124} are given as follows: Given an image, U2-Net ~\cite{U2-Net} is first used to extract the foreground. Next, we utilize LGM ~\cite{LGM} as our model to generate static 3D Gaussians from a single image, with LGM settings consistent with those in the original implementation. For the 4D dynamics generation phase, we employ MPM to generate dynamics, with the MPM settings following those in ~\cite{Physgaussian}. More details settings are provided in the appendix.

\subsection{Comparisons with State-of-the-Art Methods}
\begin{figure*}[htbp]
	\centering
	\includegraphics[width=1\linewidth]{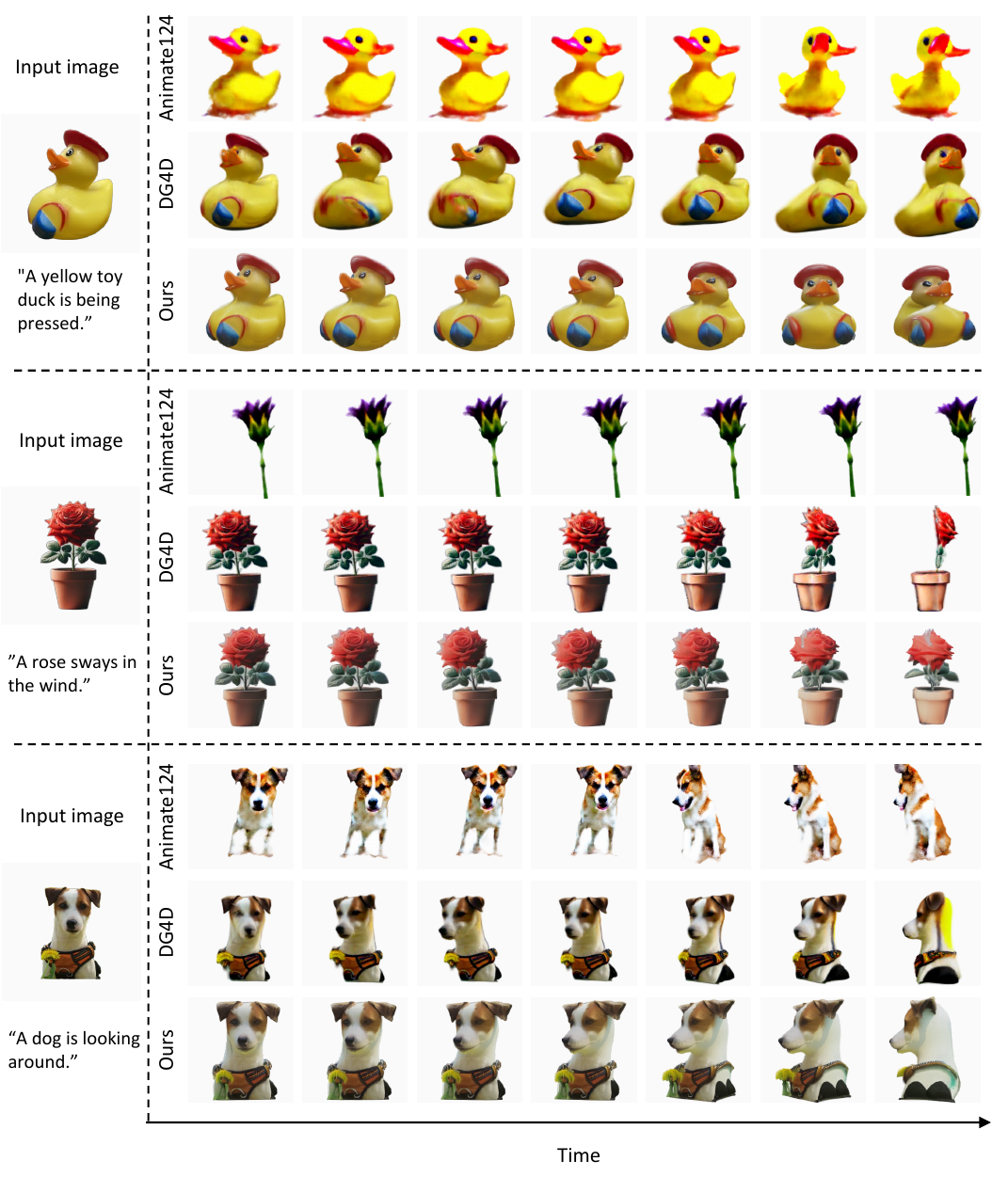}
	\caption{Qualitative comparison with the baseline methods for image-to-4D generation. \textbf{The description below the input image outlines the 4D content the user aims to generate}. For each method, 14 frames of 4D content are generated, and every second frame is selected for display, showing a total of seven frames. Additionally, to compare geometric and temporal consistency across multiple views, the rendering perspective will change with each time step.
    }
	\label{fig:main_exp}
\end{figure*}


\begin{figure*}[htbp]
	\centering
	\includegraphics[width=1\linewidth]{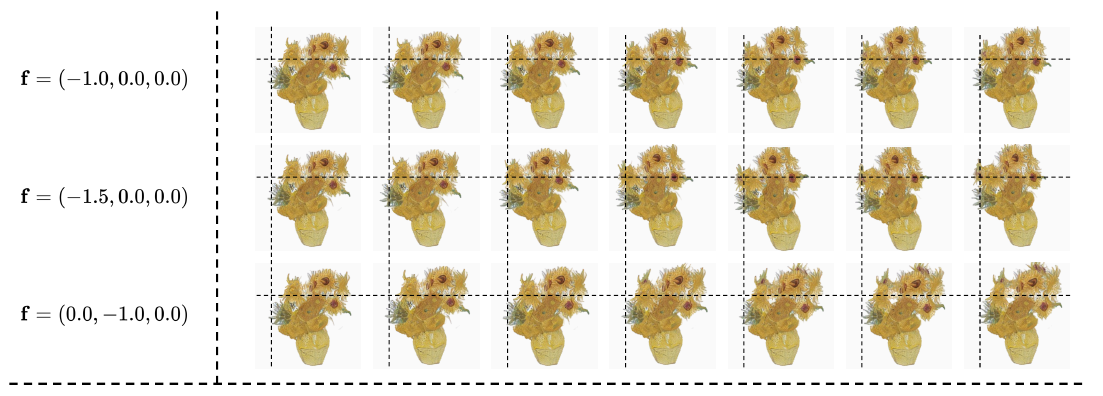}
	\caption{4D content generated by applying different external forces. $\mathbf{f}$ denotes the external forces applied along the $x$, $y$, and $z$ directions. Dashed lines are included in the images to help observe the motion.}
	\label{fig:different_force_exp}
\end{figure*}

\begin{figure}[htbp]
	\centering
	\includegraphics[scale=0.88]{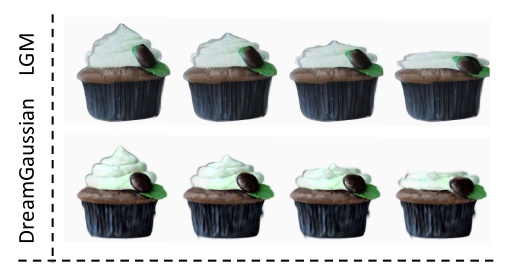}
	\caption{4D content generated using different 3D generation methods. 
    }
	\label{fig:different_module_exp}
\end{figure}

\subsubsection{Qualitative Results.}
Fig. \ref{fig:main_exp} shows the qualitative comparison between our method and other methods. 
The text description below the input image specifies the desired dynamic effect. For example, a yellow toy duck is being pressed. Fig. \ref{fig:main_exp} clearly shows that our generated 4D content exhibits high fidelity and outperforms the baseline methods. The results indicate that: Firstly, \textit{Phy124} is capable of generating 4D content that conforms to physical laws, which is largely due to its physics-driven approach. Additionally, \textit{Phy124} offers controllability, allowing the generated 4D content to align well with the user’s intent. Animate124 struggles to generate 4D content that aligns with the input image, and its generated 4D content is almost entirely static. The process of generating 4D content with DreamGaussian4D heavily depends on the quality of the reference video. However, it is challenging to obtain a reference video from existing video diffusion models that both adhere to physical laws and align with user intent. As a result, the 4D content generated by DreamGaussian4D is hard to control and inconsistent with physical laws. To compare the spatiotemporal consistency, the rendering view changes with each time step. The figures illustrate that \textit{Phy124} generates 4D content with better geometric and temporal consistency than the baseline methods.

\subsubsection{Quantitative Results.} Table \ref{tab:main_exp} provides the quantitative comparison between our method and other approaches. We evaluate both video quality using the CLIP-T score~\cite{CLIP} and generation time. 
To further quantify spatial and temporal consistency, we rendered videos from the front, right, back, and left views and computed CLIP-T score for each view. Table \ref{tab:main_exp} shows that our approach outperforms the baseline methods across all quantitative metrics. This indicates that \textit{Phy124} generates 4D content that conforms to physical laws, and ensures spatial and temporal consistency.
Meanwhile, \textit{Phy124} requires only 39.5 seconds to generate 4D content, which is significantly less time than the baseline methods. 

\subsection{Ablation Studies and Analysis}
\subsubsection{The Ability for Controlling 4D Dynamics.}
Fig. \ref{fig:different_force_exp} illustrates the 4D content generated from various external forces, demonstrating the controllability of \textit{Phy124}. In the first row, external force $f=(-1.0,0.0,0.0)$, directed to the right along the x-axis is applied, causing the sunflower to move to the right. In the second row, a larger force $f=(-1.5,0.0,0.0)$ is applied in the same direction, resulting in more intense motion, as shown in Fig. \ref{fig:different_force_exp}. This demonstrates that \textit{Phy124} can control the strength of the motion by adjusting the magnitude of the external force. In the third row, a forward force $f=(0.0,-1.0,0.0)$  of the same magnitude as in the first row is applied along the y-axis.
As a result, the sunflower moves forward under this external force.
This indicates that \textit{Phy124} can control the direction of the motion by altering the direction of the external force. To summarize, \textit{Phy124} can manage dynamics in 4D content using external forces to align with user intentions.

\subsubsection{Plug-and-Play Capability.}
\textit{Phy124} clearly separates the 4D generation process into two distinct phases: 3D Gaussians generation and 4D dynamics generation. This design allows our framework to be independent of any specific 3DGS-based 3D generation method, granting \textit{Phy124} plug-and-play capability. This provides our framework with greater flexibility and applicability.
To explore the plug-and-play capability of \textit{Phy124}, we select various 3D generation methods, including LGM ~\cite{LGM} and DreamGaussian ~\cite{DreamGaussian}.
As shown in Fig. \ref{fig:different_module_exp}, both methods generate similar dynamics and motion, which adhere to physical laws and remain controllable.
This demonstrates that our approach, regardless of the specific 3D generation methods used, exhibits strong plug-and-play capabilities.
As a result, while superior 3D generation techniques enhance the visual quality of the 4D content produced by \textit{Phy124}, they do not affect the dynamics or motion.
\section{Conclusion}
In this paper, we have introduced \textit{Phy124}, a novel, fast, physics-driven framework for generating 4D content from a single image. By integrating physical simulation directly into the 4D generation process, \textit{Phy124} ensures that the generated 4D content adheres to natural physical laws. To achieve controllable 4D generation, \textit{Phy124} incorporates external forces, allowing precise manipulation of the dynamics, such as movement speed and direction, to align with user intentions. Furthermore, by eliminating the time-consuming score distillation sampling phase, \textit{Phy124} significantly reduces the time required for 4D content generation.
Our extensive experiments demonstrate that \textit{Phy124} not only generates physically accurate and high-fidelity 4D content but also achieves this with markedly reduced inference times, pushing a new boundary in the field.


\begin{thebibliography}{10}\itemsep=-1pt

\bibitem{4D-FY}
Sherwin Bahmani, Ivan Skorokhodov, Victor Rong, Gordon Wetzstein, Leonidas Guibas, Peter Wonka, Sergey Tulyakov, Jeong~Joon Park, Andrea Tagliasacchi, and David~B Lindell.
\newblock 4d-fy: Text-to-4d generation using hybrid score distillation sampling.
\newblock In {\em Proceedings of the IEEE/CVF Conference on Computer Vision and Pattern Recognition}, pages 7996--8006, 2024.

\bibitem{Art_Create}
Jason Bailey.
\newblock The tools of generative art, from flash to neural networks.
\newblock {\em Art in America}, 8:1, 2020.

\bibitem{Mip-NeRF}
Jonathan~T Barron, Ben Mildenhall, Matthew Tancik, Peter Hedman, Ricardo Martin-Brualla, and Pratul~P Srinivasan.
\newblock Mip-nerf: A multiscale representation for anti-aliasing neural radiance fields.
\newblock In {\em Proceedings of the IEEE/CVF international conference on computer vision}, pages 5855--5864, 2021.

\bibitem{DM-video-2}
Andreas Blattmann, Tim Dockhorn, Sumith Kulal, Daniel Mendelevitch, Maciej Kilian, Dominik Lorenz, Yam Levi, Zion English, Vikram Voleti, Adam Letts, et~al.
\newblock Stable video diffusion: Scaling latent video diffusion models to large datasets.
\newblock {\em arXiv preprint arXiv:2311.15127}, 2023.

\bibitem{Hexplane}
Ang Cao and Justin Johnson.
\newblock Hexplane: A fast representation for dynamic scenes.
\newblock In {\em Proceedings of the IEEE/CVF Conference on Computer Vision and Pattern Recognition}, pages 130--141, 2023.

\bibitem{3DGeneration-text-1}
Rui Chen, Yongwei Chen, Ningxin Jiao, and Kui Jia.
\newblock Fantasia3d: Disentangling geometry and appearance for high-quality text-to-3d content creation.
\newblock In {\em Proceedings of the IEEE/CVF international conference on computer vision}, pages 22246--22256, 2023.

\bibitem{Dreamscene4d}
Wen-Hsuan Chu, Lei Ke, and Katerina Fragkiadaki.
\newblock Dreamscene4d: Dynamic multi-object scene generation from monocular videos.
\newblock {\em arXiv preprint arXiv:2405.02280}, 2024.

\bibitem{MPM-4}
Gilles Daviet and Florence Bertails-Descoubes.
\newblock A semi-implicit material point method for the continuum simulation of granular materials.
\newblock {\em ACM Transactions on Graphics (TOG)}, 35(4):1--13, 2016.

\bibitem{Gaussianflow}
Quankai Gao, Qiangeng Xu, Zhe Cao, Ben Mildenhall, Wenchao Ma, Le Chen, Danhang Tang, and Ulrich Neumann.
\newblock Gaussianflow: Splatting gaussian dynamics for 4d content creation.
\newblock {\em arXiv preprint arXiv:2403.12365}, 2024.

\bibitem{gong2024cross}
Yunpeng Gong et~al.
\newblock Cross-modality perturbation synergy attack for person re-identification.
\newblock {\em arXiv preprint arXiv:2401.10090}, 2024.

\bibitem{gong2024beyond}
Yunpeng Gong, Yongjie Hou, Chuangliang Zhang, and Min Jiang.
\newblock Beyond augmentation: Empowering model robustness under extreme capture environments.
\newblock {\em arXiv preprint arXiv:2407.13640}, 2024.

\bibitem{gong2021eliminate}
Yunpeng Gong, Liqing Huang, and Lifei Chen.
\newblock Eliminate deviation with deviation for data augmentation and a general multi-modal data learning method.
\newblock {\em arXiv preprint arXiv:2101.08533}, 2021.

\bibitem{gong2022person}
Yunpeng Gong, Liqing Huang, and Lifei Chen.
\newblock Person re-identification method based on color attack and joint defence.
\newblock In {\em Proceedings of the IEEE/CVF conference on computer vision and pattern recognition}, pages 4313--4322, 2022.

\bibitem{gong2024exploring}
Yunpeng Gong, Jiaquan Li, Lifei Chen, and Min Jiang.
\newblock Exploring color invariance through image-level ensemble learning.
\newblock {\em arXiv preprint arXiv:2401.10512}, 2024.

\bibitem{gong2024beyond2}
Yunpeng Gong, Chuangliang Zhang, Yongjie Hou, Lifei Chen, and Min Jiang.
\newblock Beyond dropout: Robust convolutional neural networks based on local feature masking.
\newblock {\em arXiv preprint arXiv:2407.13646}, 2024.

\bibitem{DM}
Jonathan Ho, Ajay Jain, and Pieter Abbeel.
\newblock Denoising diffusion probabilistic models.
\newblock {\em Advances in neural information processing systems}, 33:6840--6851, 2020.

\bibitem{3DGeneration-text-2}
Yukun Huang, Jianan Wang, Yukai Shi, Xianbiao Qi, Zheng-Jun Zha, and Lei Zhang.
\newblock Dreamtime: An improved optimization strategy for text-to-3d content creation.
\newblock {\em arXiv preprint arXiv:2306.12422}, 2023.

\bibitem{MPM-2}
Chenfanfu Jiang, Craig Schroeder, Joseph Teran, Alexey Stomakhin, and Andrew Selle.
\newblock The material point method for simulating continuum materials.
\newblock In {\em Acm siggraph 2016 courses}, pages 1--52. 2016.

\bibitem{Consistent4D}
Yanqin Jiang, Li Zhang, Jin Gao, Weimin Hu, and Yao Yao.
\newblock Consistent4d: Consistent 360 $\{$$\backslash$deg$\}$ dynamic object generation from monocular video.
\newblock {\em arXiv preprint arXiv:2311.02848}, 2023.

\bibitem{Shape-e}
Heewoo Jun and Alex Nichol.
\newblock Shap-e: Generating conditional 3d implicit functions.
\newblock {\em arXiv preprint arXiv:2305.02463}, 2023.

\bibitem{3DGS}
Bernhard Kerbl, Georgios Kopanas, Thomas Leimk{\"u}hler, and George Drettakis.
\newblock 3d gaussian splatting for real-time radiance field rendering.
\newblock {\em ACM Trans. Graph.}, 42(4):139--1, 2023.

\bibitem{3DGeneration-text-4}
Yuhan Li, Yishun Dou, Yue Shi, Yu Lei, Xuanhong Chen, Yi Zhang, Peng Zhou, and Bingbing Ni.
\newblock Focaldreamer: Text-driven 3d editing via focal-fusion assembly.
\newblock In {\em Proceedings of the AAAI Conference on Artificial Intelligence}, volume~38, pages 3279--3287, 2024.

\bibitem{Diffusion4D}
Hanwen Liang, Yuyang Yin, Dejia Xu, Hanxue Liang, Zhangyang Wang, Konstantinos~N Plataniotis, Yao Zhao, and Yunchao Wei.
\newblock Diffusion4d: Fast spatial-temporal consistent 4d generation via video diffusion models.
\newblock {\em arXiv preprint arXiv:2405.16645}, 2024.

\bibitem{One2345}
Minghua Liu, Chao Xu, Haian Jin, Linghao Chen, Mukund Varma~T, Zexiang Xu, and Hao Su.
\newblock One-2-3-45: Any single image to 3d mesh in 45 seconds without per-shape optimization.
\newblock {\em Advances in Neural Information Processing Systems}, 36, 2024.

\bibitem{Zero123}
Ruoshi Liu, Rundi Wu, Basile Van~Hoorick, Pavel Tokmakov, Sergey Zakharov, and Carl Vondrick.
\newblock Zero-1-to-3: Zero-shot one image to 3d object.
\newblock In {\em Proceedings of the IEEE/CVF international conference on computer vision}, pages 9298--9309, 2023.

\bibitem{SyncDreamer}
Yuan Liu, Cheng Lin, Zijiao Zeng, Xiaoxiao Long, Lingjie Liu, Taku Komura, and Wenping Wang.
\newblock Syncdreamer: Generating multiview-consistent images from a single-view image.
\newblock {\em arXiv preprint arXiv:2309.03453}, 2023.

\bibitem{Wonder3d}
Xiaoxiao Long, Yuan-Chen Guo, Cheng Lin, Yuan Liu, Zhiyang Dou, Lingjie Liu, Yuexin Ma, Song-Hai Zhang, Marc Habermann, Christian Theobalt, et~al.
\newblock Wonder3d: Single image to 3d using cross-domain diffusion.
\newblock In {\em Proceedings of the IEEE/CVF Conference on Computer Vision and Pattern Recognition}, pages 9970--9980, 2024.

\bibitem{3DGeneration-image-1}
Luke Melas-Kyriazi, Iro Laina, Christian Rupprecht, and Andrea Vedaldi.
\newblock Realfusion: 360deg reconstruction of any object from a single image.
\newblock In {\em Proceedings of the IEEE/CVF conference on computer vision and pattern recognition}, pages 8446--8455, 2023.

\bibitem{NeRF}
Ben Mildenhall, Pratul~P Srinivasan, Matthew Tancik, Jonathan~T Barron, Ravi Ramamoorthi, and Ren Ng.
\newblock Nerf: Representing scenes as neural radiance fields for view synthesis.
\newblock {\em Communications of the ACM}, 65(1):99--106, 2021.

\bibitem{Instant-NeRF}
Thomas M{\"u}ller, Alex Evans, Christoph Schied, and Alexander Keller.
\newblock Instant neural graphics primitives with a multiresolution hash encoding.
\newblock {\em ACM transactions on graphics (TOG)}, 41(4):1--15, 2022.

\bibitem{Point-e}
Alex Nichol, Heewoo Jun, Prafulla Dhariwal, Pamela Mishkin, and Mark Chen.
\newblock Point-e: A system for generating 3d point clouds from complex prompts.
\newblock {\em arXiv preprint arXiv:2212.08751}, 2022.

\bibitem{Efficient4D}
Zijie Pan, Zeyu Yang, Xiatian Zhu, and Li Zhang.
\newblock Fast dynamic 3d object generation from a single-view video.
\newblock {\em arXiv preprint arXiv:2401.08742}, 2024.

\bibitem{DreamFusion}
Ben Poole, Ajay Jain, Jonathan~T Barron, and Ben Mildenhall.
\newblock Dreamfusion: Text-to-3d using 2d diffusion.
\newblock {\em arXiv preprint arXiv:2209.14988}, 2022.

\bibitem{3DGeneration-image-2}
Guocheng Qian, Jinjie Mai, Abdullah Hamdi, Jian Ren, Aliaksandr Siarohin, Bing Li, Hsin-Ying Lee, Ivan Skorokhodov, Peter Wonka, Sergey Tulyakov, et~al.
\newblock Magic123: One image to high-quality 3d object generation using both 2d and 3d diffusion priors.
\newblock {\em arXiv preprint arXiv:2306.17843}, 2023.

\bibitem{U2-Net}
Xuebin Qin, Zichen Zhang, Chenyang Huang, Masood Dehghan, Osmar~R Zaiane, and Martin Jagersand.
\newblock U2-net: Going deeper with nested u-structure for salient object detection.
\newblock {\em Pattern recognition}, 106:107404, 2020.

\bibitem{CLIP}
Alec Radford, Jong~Wook Kim, Chris Hallacy, Aditya Ramesh, Gabriel Goh, Sandhini Agarwal, Girish Sastry, Amanda Askell, Pamela Mishkin, Jack Clark, et~al.
\newblock Learning transferable visual models from natural language supervision.
\newblock In {\em International conference on machine learning}, pages 8748--8763. PMLR, 2021.

\bibitem{MPM-3}
Daniel Ram, Theodore Gast, Chenfanfu Jiang, Craig Schroeder, Alexey Stomakhin, Joseph Teran, and Pirouz Kavehpour.
\newblock A material point method for viscoelastic fluids, foams and sponges.
\newblock In {\em Proceedings of the 14th ACM SIGGRAPH/Eurographics Symposium on Computer Animation}, pages 157--163, 2015.

\bibitem{DreamGaussian4D}
Jiawei Ren, Liang Pan, Jiaxiang Tang, Chi Zhang, Ang Cao, Gang Zeng, and Ziwei Liu.
\newblock Dreamgaussian4d: Generative 4d gaussian splatting.
\newblock {\em arXiv preprint arXiv:2312.17142}, 2023.

\bibitem{DM-image-1}
Robin Rombach, Andreas Blattmann, Dominik Lorenz, Patrick Esser, and Bj{\"o}rn Ommer.
\newblock High-resolution image synthesis with latent diffusion models.
\newblock In {\em Proceedings of the IEEE/CVF conference on computer vision and pattern recognition}, pages 10684--10695, 2022.

\bibitem{3DGeneration-text-5}
Junyoung Seo, Wooseok Jang, Min-Seop Kwak, Hyeonsu Kim, Jaehoon Ko, Junho Kim, Jin-Hwa Kim, Jiyoung Lee, and Seungryong Kim.
\newblock Let 2d diffusion model know 3d-consistency for robust text-to-3d generation.
\newblock {\em arXiv preprint arXiv:2303.07937}, 2023.

\bibitem{MVDream}
Yichun Shi, Peng Wang, Jianglong Ye, Mai Long, Kejie Li, and Xiao Yang.
\newblock Mvdream: Multi-view diffusion for 3d generation.
\newblock {\em arXiv preprint arXiv:2308.16512}, 2023.

\bibitem{MAV3D}
Uriel Singer, Shelly Sheynin, Adam Polyak, Oron Ashual, Iurii Makarov, Filippos Kokkinos, Naman Goyal, Andrea Vedaldi, Devi Parikh, Justin Johnson, et~al.
\newblock Text-to-4d dynamic scene generation.
\newblock {\em arXiv preprint arXiv:2301.11280}, 2023.

\bibitem{DM-implicit}
Jiaming Song, Chenlin Meng, and Stefano Ermon.
\newblock Denoising diffusion implicit models.
\newblock {\em arXiv preprint arXiv:2010.02502}, 2020.

\bibitem{MPM-1}
Alexey Stomakhin, Craig Schroeder, Lawrence Chai, Joseph Teran, and Andrew Selle.
\newblock A material point method for snow simulation.
\newblock {\em ACM Transactions on Graphics (TOG)}, 32(4):1--10, 2013.

\bibitem{EG4D}
Qi Sun, Zhiyang Guo, Ziyu Wan, Jing~Nathan Yan, Shengming Yin, Wengang Zhou, Jing Liao, and Houqiang Li.
\newblock Eg4d: Explicit generation of 4d object without score distillation.
\newblock {\em arXiv preprint arXiv:2405.18132}, 2024.

\bibitem{LGM}
Jiaxiang Tang, Zhaoxi Chen, Xiaokang Chen, Tengfei Wang, Gang Zeng, and Ziwei Liu.
\newblock Lgm: Large multi-view gaussian model for high-resolution 3d content creation.
\newblock {\em arXiv preprint arXiv:2402.05054}, 2024.

\bibitem{DreamGaussian}
Jiaxiang Tang, Jiawei Ren, Hang Zhou, Ziwei Liu, and Gang Zeng.
\newblock Dreamgaussian: Generative gaussian splatting for efficient 3d content creation.
\newblock {\em arXiv preprint arXiv:2309.16653}, 2023.

\bibitem{3DGeneration-image-3}
Junshu Tang, Tengfei Wang, Bo Zhang, Ting Zhang, Ran Yi, Lizhuang Ma, and Dong Chen.
\newblock Make-it-3d: High-fidelity 3d creation from a single image with diffusion prior.
\newblock In {\em Proceedings of the IEEE/CVF international conference on computer vision}, pages 22819--22829, 2023.

\bibitem{ImageDream}
Peng Wang and Yichun Shi.
\newblock Imagedream: Image-prompt multi-view diffusion for 3d generation.
\newblock {\em arXiv preprint arXiv:2312.02201}, 2023.

\bibitem{DM-video-1}
Yaohui Wang, Xinyuan Chen, Xin Ma, Shangchen Zhou, Ziqi Huang, Yi Wang, Ceyuan Yang, Yinan He, Jiashuo Yu, Peiqing Yang, et~al.
\newblock Lavie: High-quality video generation with cascaded latent diffusion models.
\newblock {\em arXiv preprint arXiv:2309.15103}, 2023.

\bibitem{3DGeneration-text-3}
Zhengyi Wang, Cheng Lu, Yikai Wang, Fan Bao, Chongxuan Li, Hang Su, and Jun Zhu.
\newblock Prolificdreamer: High-fidelity and diverse text-to-3d generation with variational score distillation.
\newblock {\em Advances in Neural Information Processing Systems}, 36, 2024.

\bibitem{Physgaussian}
Tianyi Xie, Zeshun Zong, Yuxing Qiu, Xuan Li, Yutao Feng, Yin Yang, and Chenfanfu Jiang.
\newblock Physgaussian: Physics-integrated 3d gaussians for generative dynamics.
\newblock In {\em Proceedings of the IEEE/CVF Conference on Computer Vision and Pattern Recognition}, pages 4389--4398, 2024.

\bibitem{Deformable3DGS}
Ziyi Yang, Xinyu Gao, Wen Zhou, Shaohui Jiao, Yuqing Zhang, and Xiaogang Jin.
\newblock Deformable 3d gaussians for high-fidelity monocular dynamic scene reconstruction.
\newblock In {\em Proceedings of the IEEE/CVF Conference on Computer Vision and Pattern Recognition}, pages 20331--20341, 2024.

\bibitem{4DGen}
Yuyang Yin, Dejia Xu, Zhangyang Wang, Yao Zhao, and Yunchao Wei.
\newblock 4dgen: Grounded 4d content generation with spatial-temporal consistency.
\newblock {\em arXiv preprint arXiv:2312.17225}, 2023.

\bibitem{DM-image-2}
Lvmin Zhang, Anyi Rao, and Maneesh Agrawala.
\newblock Adding conditional control to text-to-image diffusion models.
\newblock In {\em Proceedings of the IEEE/CVF International Conference on Computer Vision}, pages 3836--3847, 2023.

\bibitem{Animate124}
Yuyang Zhao, Zhiwen Yan, Enze Xie, Lanqing Hong, Zhenguo Li, and Gim~Hee Lee.
\newblock Animate124: Animating one image to 4d dynamic scene.
\newblock {\em arXiv preprint arXiv:2311.14603}, 2023.

\bibitem{Nvidia}
Yufeng Zheng, Xueting Li, Koki Nagano, Sifei Liu, Otmar Hilliges, and Shalini De~Mello.
\newblock A unified approach for text-and image-guided 4d scene generation.
\newblock In {\em Proceedings of the IEEE/CVF Conference on Computer Vision and Pattern Recognition}, pages 7300--7309, 2024.

\end{thebibliography}

\end{document}